\documentclass[preprint,1p,11pt]{ISAS_IR}

\usepackage[english]{babel}
\usepackage[latin1]{inputenc}
\usepackage{calc}
\usepackage{amsmath}
\usepackage{amssymb}
\usepackage{microtype}
\usepackage{lmodern}
\usepackage{amsmath} 
\usepackage{amssymb}  
\usepackage{mathMakros}
\usepackage[colorinlistoftodos]{todonotes}
\usepackage{algorithm}
\usepackage{algorithmic}
\usepackage{tabularx}
\usepackage{subfig}
\usepackage{siunitx}
\usepackage{mathtools}
\mathtoolsset{showonlyrefs}

\input{Defs.h} 

\begin{document}

\begin{frontmatter}

\title{Proximal Policy Optimization for Tracking Control \\ Exploiting Future Reference Information}

\author[isas]{Jana~Mayer}
\ead{jana.mayer@kit.edu}

\author[isas]{Johannes~Westermann}
\ead{johannes.westermann@kit.edu}

\author[isas]{Juan~Pedro~Guti\'{e}rrez~H.~Muriedas}
\ead{juanpedroghm@gmail.com}

\author[iav]{Uwe~Mettin}
\ead{uwe.mettin@iav.de}

\author[iav]{Alexander~Lampe}
\ead{alexander.lampe@iav.de}

\address[isas]{Intelligent Sensor-Actuator-Systems Laboratory (ISAS)\\
Institute for Anthropomatics and Robotics\\
Karlsruhe Institute of Technology (KIT), Germany}

\address[iav]{IAV GmbH, Berlin, Germany}

\begin{abstract}
In recent years, reinforcement learning (RL) has gained increasing attention in control engineering. Especially, policy gradient methods are widely used. 
In this work, we improve the tracking performance of proximal policy optimization (PPO) for arbitrary reference signals by incorporating information about future reference values. 
Two variants of extending  the argument of the actor and the critic taking future reference values into account are presented. In the first variant, global future reference values are added to the argument. For the second variant, a novel kind of residual space with future reference values applicable to model-free reinforcement learning is introduced. Our approach is evaluated against a PI controller on a simple drive train model. We expect our method to generalize to arbitrary references better than previous approaches, pointing towards the applicability of RL to control real systems. 
\end{abstract}

\end{frontmatter}

%
%
\section{Introduction}
In cars with automatic transmissions, gear shifts shall be performed in such a way that no discomfort is caused by the interplay of motor torque and clutch control action. Especially, the synchronization of input speed to the gears' target speed is a sensitive process mainly controlled by the clutch. Ideally, the motor speed should follow a predesigned reference, but not optimal operating feedforward control and disturbances in the mechanical system can cause deviations from the optimal behavior. The idea is to apply a reinforcement learning (RL) approach to control the clutch behavior regulating the deviations from the optimal reference. The advantage of a RL approach over classical approaches as the PI control is that no extensive experimental parametrization for every gear and every clutch is needed which is very complex for automatic transmissions. Instead, the RL algorithm is supposed to learn the optimal control behavior autonomously. 
Since the goal is to guide the motor speed along a given reference signal the problem at hand belongs to the family of tracking control problems. 

In the following, a literature review on RL for tracking control is given. 
In \cite{qin_2018}, the deep deterministic policy gradient method, introduced in \cite{lillicrap_2016}, is applied to learn the parameters of a  PID controller. An adaptive PID controller is realized in \cite{carlucho_2017} using an incremental Q-learning for real-time tuning. A combination of  Q-learning and a PID controller is presented in \cite{wang_2020}, where the applied control input is a sum of the PID control input and a control input determined by Q-learning. 

Another common concept applied to tracking control problems is model predictive control (MPC) which can also be combined with RL. A data-efficient model-based RL approach based on probabilistic model predictive control (MPC) is introduced in \cite{kamthe_2018}. The key idea is to learn the probabilistic transition model using Gaussian processes. In \cite{gros_2020}, nonlinear model predictive control (NMPC) is used as a function approximator for the value function and the policy in a RL approach. For tracking control problems with linear dynamics and quadratic costs a RL approach  is presented in \cite{koepf_2020}. Here, a Q-function is analytically derived that inherently incorporates a given reference trajectory on a moving horizon.

In contrast to the before presented approaches derived from classical controller concepts also pure 
RL approaches for tracking control were invented. A model-based variant is presented in \cite{hu_2020} where a kernel-based transition dynamic model is introduced. The transition probabilities are learned directly from the observed data without learning the dynamic model. The model of the transition probabilities is then used in a RL approach.

A model-free RL approach is introduced in \cite{Yu_2017} where the deep deterministic policy gradient approach \cite{lillicrap_2016} is applied on tracking control of an autonomous underwater vehicle. In  \cite{kamran_2019}, images of a reference are fed to a convolutional neural network for a model-free state representation of the path. A deep deterministic policy gradient approach \cite{lillicrap_2016} is applied where previous local path images and control inputs are given as arguments to solve the tracking control problem. 
Proximal policy optimization (PPO) \cite{schulman_2017} with generalized advantage estimation (GAE) \cite{schulman_2016} is applied on tracking control of a manipulator and a mobile robot in \cite{zhang_2019}. Here, the actor and the critic are represented by a long short-term memory (LSTM) and a distributed version of PPO is used.

In this work, we apply PPO to a tracking control problem. The key idea is to extend the arguments of the actor and the critic to take into account information about future reference values and thus improve the tracking performance. Besides adding global reference values to the argument, we also define an argument based on residua between the states and the future reference values. For this purpose, a novel residual space with future reference values is introduced applicable to model-free RL approaches. Our approach is evaluated on a simple drive train model. The results are compared to a classical PI controller and a PPO approach, which does not consider future reference values.

\section{Problem Formulation}
In this work, we consider a time-discrete system with non-linear dynamics
\begin{equation}
    \Vec{x}_{k+1} = f \left( \Vec{x}_k, \Vec{u}_k\right) \enspace ,
\label{eq:sy}   
\end{equation}
where $\vec{x}_k \in \mathbb{R}^{n_x}$ is the state and $\Vec{u}_k \in \mathbb{R}^{n_u}$ is the control input applied in time step $k$.
The system equation \eqref{eq:sy} is assumed to be unknown for the RL algorithm. Furthermore, the states $\vec{x}_k$ are exactly known.

In the tracking control problem, the state or components of the state are supposed to follow a reference $\vec{x}^r_k \in \mathbb{R}^{n_h}$. 
Thus, the goal is to control the system in a way that the deviation between the state $\vec{x}_k$ and the reference $\vec{x}^r_k$ becomes zero in all time steps. The reference is assumed to be given and the algorithm should be able to track before unseen references.

To reach the goal, the algorithm can learn from interactions with the system. 
In policy gradient methods, a policy is determined which maps the RL states~$\vec{s}_k$ to a control input~$\vec{u}_k$. The states~$\vec{s}_k$ can be the system states~$\vec{x}_k$ but can also contain other related components. In actor-critic approaches, the policy is represented by the actor. Here, $\vec{s}_k$ is the argument given to the actor and in most instances also to the critic as input. To prevent confusion with the system state $\vec{x}_k$, we will refer to $\vec{s}_k$ as argument in the following.

\section{Existing solutions and challenges}
In existing policy gradient methods for tracking control, the argument $\vec{s}_k$ is either identical to the system state $\vec{x}_k$~\cite{Yu_2017} or 
is composed of the system state and the residuum between the system state and the reference in the current time step~\cite{zhang_2019}. Those approaches show good results if the reference is fixed. Applied on arbitrary references the actor can only respond to the current reference value but is not able to act optimally for the subsequent reference values. In this work, we will show that this results in poor performance.

Another common concept, applied on tracking control problems, is model predictive control (MPC), e.g., \cite{kamthe_2018}. Here, the control inputs are determined by predicting the future system states and minimizing their deviation from the future reference values. In general, the optimization over a moving horizon has to be executed in every time step as no explicit policy representation is determined. Another disadvantage of MPC is the need to know or learn the model of the system.

Our idea is to transfer the concept of utilizing the information given in form of known future reference values from MPC to policy gradient methods. A first step in this direction is presented in \cite{koepf_2020}, where an adaptive optimal control method for reference tracking was developed. Here, a Q-function could be analytically derived by applying dynamic programming. The received Q-function depends inherently on the current state but also on current and future reference values. However, the analytical solution is limited to the case of linear system dynamics and quadratic costs (rewards). In this work, we transfer those results 
to a policy gradient algorithm by extending the arguments of the actor and  the critic with future reference values. In contrast to the linear quadratic case, the Q-function cannot be derived analytically. Accordingly, the nonlinear dependencies are approximated by the actor and the critic. The developed approach can be applied to tracking control problems with nonlinear system dynamics and arbitrary references. 
In some applications, the local deviation of the state from the reference is more informative than the global state and reference values, e.g. operating the drive train in different speed ranges. In this case, it can be beneficial if the argument is defined as residuum between the state and the corresponding reference value, because this scales down the range of the state space has to be explored. Thus, we introduce a novel kind of residual space between states and future reference values which can be applied without knowing or learning the system dynamics. The key ideas are
(1)~extending the arguments of the actor and the critic of a policy gradient method by future reference values, and
(2)~introducing a novel kind of residual space for model-free RL.


\section{Preliminaries: Proximal Policy Optimization Algorithm}
\label{sec:PPO}
As policy gradient method, proximal policy optimization (PPO) \cite{schulman_2017} is applied in this work. PPO is a simplification of the trust region policy optimization (TRPO) \cite{schulman_2017_2}. The key idea of PPO is a novel loss function design where the change of the stochastic policy $\pi_{\vec{\theta}}$ in each update step is limited introducing a clip function
\begin{equation}
    J (\vec{\theta}) = \mathbb{E}_{\left(\vec{s}_k, \vec{u}_k\right)} \lbrace \min \left(p_k (\vec{\theta}) A_k, \text{clip} \left( p_k(\vec{\theta}),1-c,1+c\right) A_k \right) \rbrace \enspace, 
\label{eq:actorloss}    
\end{equation} 
where
\begin{equation}
    p_k(\vec{\theta}) = \frac{\pi_{\vec{\theta}} (\vec{u}_k \vert \vec{s}_k) }{\pi_{\vec{\theta}_{old}} (\vec{u}_k \vert \vec{s}_k)} \enspace .
\label{eq:pPPO}    
\end{equation}
The clipping motivates $p_k(\vec{\theta})$ not to leave the interval $[1-c,1+c]$. 
The argument $\vec{s}_k$ given to the actor commonly contains the system state $\vec{x}_k$ of the system, but can be extended by additional information.
The loss function is used in a policy gradient method to learn the actor network's parameters~$\vec{\theta}_h$
\begin{equation}
    \vec{\theta}_{h+1} = \vec{\theta}_h + \alpha_{a} \cdot \nabla_{\vec{\theta}} J (\vec{\theta}) \vert_{\vec{\theta} = \vec{\theta}_h} \enspace,
\end{equation}
where $\alpha_{a}$ is referred as the actor's learning rate and $h \in \lbrace 0,1,2, \ldots \rbrace$ is the policy update number. A proof of convergence for PPO is presented in \cite{holzleitner_2020}. 
The advantage function $A_k$ in \eqref{eq:actorloss} is defined as the difference between the Q-function and the value function $V$
\begin{equation}
    A_k(\vec{s}_k,\vec{u}_k) = Q(\vec{s}_k,\vec{u}_k) - V(\vec{s}_k) \enspace .
\end{equation}
In \cite{schulman_2017}, generalized advantage estimation (GAE) \cite{schulman_2016} is applied to approximate the value function
\begin{equation}
    \hat{A}_k^{GAE(\gamma, \lambda)} = \sum_{l = 0}^\infty \left( \gamma \lambda \right)^l \delta_{k+l}^V \enspace,
    \label{eq:adv}
\end{equation}
where $\delta_k^V$ is the temporal difference error of the value function~V \cite{sutton_2018}
\begin{equation}
    \delta_k^V = r_k + \gamma V(\vec{s}_{k+1}) -V(\vec{s}_{k}) \enspace .
\label{eq:TDE}    
\end{equation}
The discount $\gamma \in [0,1]$ reduces the influence of future incidences. $\lambda \in [0,1]$ is a design parameter of GAE. Note, the value function also has to be learned during the training process and serves as critic in the approach. The critic is also represented by a neural network and and receives the same argument as the actor.
To ensure sufficient exploration the actor's loss function \eqref{eq:actorloss} is commonly extended by an entropy bonus $S[\pi_{\vec{\theta}}] (\vec{s}_k)$ \cite{mnih_2016}, \cite{williams_1991} 
\begin{equation}
    J (\vec{\theta}) = \mathbb{E}_{(\vec{s}_k, \vec{u}_k)} \lbrace \min \left(p_k (\vec{\theta}) A_k, \text{clip} \left( p_k(\vec{\theta}),1-c,1+c\right) A_k \right) \\
    + \mu \, S[\pi_{\vec{\theta}}] (\vec{s}_k) \rbrace \enspace,
\end{equation}
where $\mu \geq 0$ is the entropy coefficient. 
\section{Proximal Policy Optimization for Tracking Control with future references}
\label{sec:PPO_TC}
As mentioned before, the key idea of the presented approach is to add the information of future reference values to the argument $\vec{s}_k$ of the actor and the critic in order to improve the control quality. 
PPO was already applied to a tracking control problem in \cite{zhang_2019} where the argument $\vec{s}_k^0$, contains the current system state $\vec{x}_k$ as well as the residuum of $\vec{x}_{k}$ and the current reference $\vec{x}^r_{k}$
\begin{equation}
    \vec{s}_k^0 = \left[ \vec{x}_k, \left(\vec{x}^r_{k} - \vec{x}_{k}\right) \right]^T \enspace .
\label{eq:arg0}
\end{equation}
However, no information of future reference values is part of the argument. We take advantage of the fact that the future reference values are known and incorporate them to the argument of the actor and the critic. In the following, two variants will be discussed:

(1) Besides the system state $\vec{x}_k$ $N$ future reference values are added to the argument
\begin{equation}
    \vec{s}_k^{1N} = \left[ \vec{x}_k, \vec{x}^r_k, \vec{x}^r_{k+1},  \ldots, \vec{x}^r_{k+N} \right]^T  , \enspace N \in \mathbb{N}  \enspace .
\label{eq:arg1}
\end{equation}

\begin{figure}
    \centering
    \def\svgwidth{0.7\linewidth}
    \graphicspath{{figures/}} 
    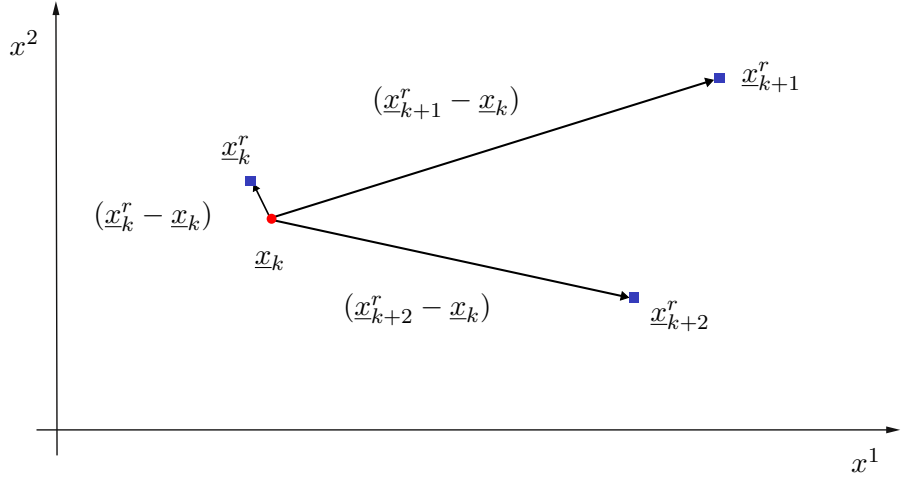 
    \caption{Residual space defined by current state and future reference values.}
    \label{fig:res}
\end{figure}

(2) We introduce a novel residual space where the future reference values are related to the current state and the argument is defined as
\begin{equation}
    \vec{s}_k^{2N} = \left[ \left(\vec{x}^r_{k} - \vec{x}_{k}\right), \left(\vec{x}^r_{k+1} -  \vec{x}_{k} \right), \ldots, \left(\vec{x}^r_{k+N} -  \vec{x}_{k}\right) \right]^T  .
\label{eq:arg2}
\end{equation} 
In Figure~\ref{fig:res}, the residual space is illustrated for two-dimensional states and references. Being in the current state~$\vec{x}_k = [x^1_k, x^2_k]^T$ (red dot) the the residua between the current state and the future reference values (black arrows) indicate if the components of the state $x^1_k$ and $x^2_k$ have to be increased or decreased to reach the reference values $\vec{x}^r_{k+1}$ and $\vec{x}^r_{k+2}$ in the next time steps. Thus, the residual argument gives sufficient information to the PPO algorithm to control the system. 

Please note, a residual space containing future states $\left(\vec{x}^r_{k+1} - \vec{x}_{k+1}\right),\ldots, \left( \vec{x}^r_{k+N} - \vec{x}_{k+N}\right)$ would suffer from two disadvantages. First,  the model has to learned which would increase the complexity of the algorithm. Second, the future states $\vec{x}_{k+1},\ldots,\vec{x}_{k+N}$ depend on the current policy thus the argument is a function of the policy. Applied as argument in~\eqref{eq:pPPO} the policy becomes a function of the policy $\pi_{\vec{\theta}} (\vec{u}_k \vert \vec{s}_k(\pi_{\vec{\theta}}))$. This could be solved by calculating the policy in a recursive manner where several instances of the policy are trained in form of different neural networks. This solution would lead to a complex optimization problem which is expected to be computationally expensive and hard to stabilize in the training. Therefore, we consider this solution as impractical. On the other hand, the residual space defined in \eqref{eq:arg2} contains all information about the future course of the reference, consequently a residual space including future states would not enhance the information content.

In the residual space, the arguments are centered around zero, independent of the actual values of the state or the reference. The advantage, compared to the argument with global reference values, is that only the deviation of the state from the reference is represented, which scales down the range of the state space has to be learned. But the residual space argument is only applicable if the change in the state for a given control input is independent of the state itself.

As part of the tracking control problem a reward function has to be designed. The reward should represent the quality of following the reference. Applying the control input $\vec{u}_k$ in the state $\vec{x}_k$ leads to the state $\vec{x}_{k+1}$ and related to this a reward depending on the difference between $\vec{x}_{k+1}$ and the reference $\vec{x}^r_{k+1}$. Additionally, a punishment of huge control inputs can be appended. The resulting reward function for time step $k$ is defined as
\begin{equation}
    r_k = - \left( \vec{x}_{k+1} - \vec{x}^r_{k+1}\right)^2 - \beta \cdot \vec{u}_{k}^2 \enspace,
\label{eq:reward}
\end{equation}
where $\beta \geq 0$ is a weighting parameter.
\section{Simple Drive Train Model}
An automatic transmission provides gear shifts characterized by two features: (1) the torque transfer, in which a clutch corresponding to the target gear takes over the drive torque, and (2) the speed synchronization, in which slip from input to output speed of the clutch is reduced such that it can be closed or controlled at low slip. In this work, we consider the reference tracking control problem for a friction clutch during synchronization phase. Its input is driven by a motor and the output is propagated through the gearbox to the wheels of the car. Speed control action can only be applied by the clutch when input and output force plates are in friction contact with slip. Generally, the aim in speed synchronization is to smoothly control the contact force of the clutch without jerking towards zero slip. This is where our RL approach for tracking control comes into account. For a smooth operation a reference to follow during the friction phase is predesigned by the developer of the drive train.

 For easier understanding, a simple drive train model is used which is derived from the motion equations of an ideal clutch \cite{quang_1998} extended  by the influence of the gearbox
\begin{align}
    J_{in} \, \dot{\omega}_{in} &= -T_{cl}+T_{in} \enspace,  \label{eq:clutchMotA}\\
    J_{out} \, \dot{\omega}_{out} &=  \theta \, T_{cl} - T_{out} \enspace ,
  \label{eq:clutchMotB}
\end{align}
where ${\omega}_{in}$ is the input speed on the motor side and ${\omega}_{out}$ is the output speed at the side of the wheels. 
Accordingly, $T_{in}$ is the  input torque and  $T_{out}$ is the output torque. The transmission ratio $\theta$ of the gearbox defines the ratio between the input and output speed. The input and output moment of inertia $J_{in}$, $J_{out}$ and the transmission ratio $\theta$ are fixed characteristics of the drive train.
The clutch is controlled varying the torque transmitted from the clutch  $T_{cl}$.

The input torque $T_{in}$ is approximated as constant while the output torque is assumed to depend linear on the output speed $T_{out} = \eta \cdot \omega_{out}$ which changes \eqref{eq:clutchMotA} and \eqref{eq:clutchMotB} to 
\begin{align}
    J_{in} \, \dot{\omega}_{in} &= -T_{cl}+T_{in} \enspace, \\
    J_{out} \, \dot{\omega}_{out} &=  \theta \, T_{cl} - \eta \cdot \omega_{out} \enspace .
  \label{eq:clutchMot2}
\end{align}
Solving the differential equations for a time interval $\Delta T$, yields the discrete system equation
\begin{equation}
    \begin{bmatrix}
    \omega_{in} \\
    \omega_{out}
    \end{bmatrix}_{k+1}
    = \bol{A}
    \begin{bmatrix}
    \omega_{in} \\
    \omega_{out}
    \end{bmatrix}_{k}
    +
    \bol{B}_1 \cdot T_{cl,k} 
    +
    \bol{B}_2 \cdot T_{in}  \enspace,
    \label{eq:systemEqCl}
\end{equation} 
where
\begin{align}
    \bol{A} &=
     \begin{bmatrix}
    1 & 0 \\
    0 & \exp \left( {-\frac{ \eta \cdot \Delta T}{J_{out}}} \right) 
    \end{bmatrix}
    \enspace, \\
    \bol{B}_1 &=
    \begin{bmatrix}
    - \frac{\Delta T}{J_{in}} \\
    \frac{\theta}{\eta} \left(1 - \exp \left( {-\frac{ \eta \cdot \Delta T}{J_{out}}} \right) \right) 
    \end{bmatrix} \, ,
    \enspace 
    \bol{B}_2= 
    \begin{bmatrix}
    \frac{\Delta T}{J_{in}} \\
    0
    \end{bmatrix} \enspace,
\end{align}
with state $\vec{x}_k = [\omega_{in} \omega_{out}]^T_k$  and the control input $u_k = T_{cl,k}$. For a friction-plate clutch, the clutch torque $T_{cl,k}$ depends on the capacity torque $T_{cap,k}$
\begin{equation}
    T_{cl,k}=T_{cap,k} \cdot \text{sign} \left(\omega_{in,k} - \theta \cdot \omega_{out,k}\right), \enspace T_{cap,k}  \geq 0 \enspace,
    \label{eq:Kupplungsmoment}
\end{equation}
which means $T_{cl,k}$ is changing its sign according to if the input speed or the output speed is higher.
The capacity torque is proportional to the contact force which is applied on the plates $T_{cap,k} \sim F_{N,k} $.

In real drive trains, the change of the control input in a time step is limited due to underlying dynamics such as pressure dynamics. For simulating this behavior, a low pass filter is applied on the control inputs. We use a first-order lag element (PT1) with the cutoff frequency $f_g$
\begin{equation}
    T'_{cl,k} =\begin{cases}
    \left(T_{cl,k} - T'_{cl,k-1} \right) \cdot \left( 1- a \right) + T'_{cl,k-1} , & \text{if $T_{cl,k} > T'_{cl,k-1}$} \enspace, \\
    \left(T'_{cl,k-1} - T_{cl,k} \right) \cdot a + T_{cl,k} , & \text{otherwise} \enspace,
    \end{cases}
\end{equation}    
where
\begin{equation}
    a = \exp { \left( - 2 \pi f_g \Delta T \right)} \enspace.
\end{equation}
In this case, the control input provided by the controller $T_{cl,k}$ is transformed to the delayed control input $T'_{cl,k}$ applied on the system.

The simple drive train model is used in three different experiments of input speed synchronization which are derived from use cases of real drive trains. In the first experiment, the references to be followed are smooth. This is associated with a usual gear shift. In real applications, delays, hidden dynamics, or change-of-mind situations can arise which make a reinitialization of the reference necessary. This behavior is modeled in the second experiment where the references contain jumps and discontinuities, respectively. Jumps in the reference can lead to fast changes in the control inputs. To still maintain a realistic behavior the lag element is applied on the control inputs. In the third experiment, we use again smooth references but in different input speed ranges. Varying drive torque demands typically cause the gear shift to start at different input speed levels.
Our approach will be evaluated on all three experiments to determine the performance for different use cases. Note, that for demonstration purposes we consider speed synchronization by clutch control input only. In real applications, clutch control is often combined with input torque feedforward.

\section{Evaluation}
As mentioned in the last chapter, we will evaluate our approach using three different experiments representing three use cases of the drive train. In every experiment, the three different arguments of the actor and the critic introduced in Chapter \ref{sec:PPO_TC} will be applied. 

For the drive train system \eqref{eq:systemEqCl}, the arguments of the actor and the critic have to be defined. The reference to be followed is only corresponding to the input speed. 
Since no reference for the output speed is given we add the output speed $\omega_{out,k}$ as global variable in all three arguments. 
Analogous to \eqref{eq:arg0}, \eqref{eq:arg1} and \eqref{eq:arg2}, the arguments of the drive train system are 
\begin{align}
    \vec{s}_k^{0,cl} &= \left[ \omega_{out,k}, \omega_{in,k}, \left( \omega^r_{in,k} - \omega_{in,k} \right) \right]^T \enspace, \\
    \vec{s}_k^{1N,cl} &= \left[\omega_{out,k}, \omega_{in,k}, \omega^r_{in,k}, \omega^r_{in,k+1}, \ldots, \omega^r_{in,k+N} \right]^T \enspace, \\
    \vec{s}_k^{2N,cl} &= \left[ \omega_{out,k}, \left(\omega^r_{in,k} - \omega_{in,k}\right), \left(\omega^r_{in,k+1} - \omega_{in,k}\right),  \ldots, \left(\omega^r_{in,k+N} - \omega_{in,k}\right) \right]^T .
\end{align}
Without future reference values as applied in \cite{zhang_2019} the argument is $\vec{s}_k^{0,cl}$, with global future reference values $\vec{s}_k^{1N,cl}$ and in residual space with future reference values $\vec{s}_k^{2N,cl}$.

The reward function for the simple drive train model derived from \eqref{eq:reward} is given as 
\begin{equation}
    r_k = -\left( \omega_{in,k+1} - {\omega}^r_{in,k+1}\right)^2 - \beta \cdot \left (T_{cl,k} - T_{in}\right)^2 \enspace.
\label{eq:rewardCl}    
\end{equation}
If $T_{cl,k} = T_{in}$ the input speed $\omega_{in}$ is not changing from time step $k$ to time step $k+1$ according to \eqref{eq:systemEqCl}. Thus, deviations of $T_{cl,k}$ from $ T_{in}$ are penalized to suppress control inputs which would cause larger changes in the state.
All parameters used in the simple drive train model are given in Table \ref{tab:paraClutch}.

\begin{table}[tb]
    \caption{Parameters of the simple drive train model.}
    \label{tab:paraClutch}
    \centering
    \begin{tabularx}{8.5cm}{p{5.0cm}|X} 
    \hline
    Parameter &  Value \tabularnewline
    \hline
    \hline
    Input moment of inertia $J_{in}$ & \SI{0.209}{kg m^2} \\
    \hline
    Output moment of inertia $J_{out}$ & \SI{86.6033}{kg m^2} \\
    \hline
    Transmission ratio $\theta$ & 10.02  \\
    \hline
    Input torque $T_{in}$ & \SI{20}{Nm} \\
    \hline
    $\eta$  & \SI{2}{(Nms)/rad} \\
    \hline 
    Time step $\Delta T$  & \SI{10}{ms}  \\
    \hline 
    \end{tabularx}
\end{table}
\subsection{Algorithm} 
\begin{algorithm}[tb!]
    \caption{PPO for tracking control }
    \label{algo:controlPPO}
    \begin{algorithmic}[1]
    \REQUIRE Replay buffer $\mathcal D$, critic parameters $\vec{\phi}$, actor parameters $\vec{\theta}$, actor learning rate $\alpha_a$, critic learning rate $\alpha_c$, target network delay factor $\tau$
    \STATE Init target critic parameters $\vec{\phi}' \leftarrow \vec{\phi}$ and $h = 0$
    \FOR {1 .. Number of episodes}
    \STATE Observe initial state $\vec{x}_0$ and new reference $\omega^r_{in}$
    \FOR {1 .. K}
    	\STATE Apply control input $u_k \leftarrow \pi_\theta(\vec{s}_k)$
    	\STATE Observe new state $\vec{x}_{k+1}$ and reward $r_k$
    	\STATE Add $(\vec{s}_k, u_k, r_k, \vec{s}_{k+1} )$ to replay buffer $\mathcal D$
    \ENDFOR	
    \FOR {1 .. Number of epochs}
    	\STATE Sample training batch from $\mathcal D$
    	\STATE Update critic $\vec{\phi}_{h+1} \leftarrow \vec{\phi}_h + \alpha_c \nabla_{\vec{\phi}} C(\vec{\phi}_h)$
    	\STATE Calculate advantage $\vec{\hat A}$ using GAE
    	\STATE Update actor $\vec{\theta}_{h+1} \leftarrow \vec{\theta}_h + \alpha_a \nabla_{\vec{\theta}} J(\vec{\theta}_h)$
    	\STATE $h \leftarrow h+1$
    	\STATE \textbf{Every m-th epoch} \\ \quad Update target critic $\vec{\phi}' \leftarrow (1 - \tau) \, \vec{\phi}' + \tau \, \vec{\phi}$
    \ENDFOR
    \ENDFOR
    \end{algorithmic}
\end{algorithm}
The PPO algorithm, applied for the evaluation, is shown in Algorithm \ref{algo:controlPPO}. We use two separate networks representing the actor and the critic. In order to improve data-efficiency we apply experience replay \cite{lin_1992}, \cite{lillicrap2015}. While interacting with the system, the tuples $(\vec{s}_k, u_k, r_k, \vec{s}_{k+1} )$ are stored in the replay buffer. In every epoch a training batch of size $L$ is sampled from the replay buffer to update the actor's and the critic's parameters. 
As introduced in Chapter \ref{sec:PPO}, the advantage function is determined via generalized advantage estimation~\eqref{eq:adv} with the GAE parameter set to $\lambda = 0$ and the discount $\gamma = 0.7$.
To improve the stability of the critic network during the training, we added a target critic network~$V'$ \cite{lillicrap2015} which is only updated every m-th epoch (in our implementation $m=2$). The critic loss is defined as the mean squared temporal difference error~\eqref{eq:TDE} 
\begin{equation}
     C(\vec{\phi}) = \frac{1}{L} \sum^L \left( r_k + \gamma \, V'(\vec{s}_{k+1}) - V(\vec{s}_{k})\right)^2 \enspace .
\end{equation}
The output of the actor provides $T_{cap}$  in \eqref{eq:Kupplungsmoment} and the sign for $T_{cl}$ is calculated through the input and output speed. To ensure $T_{cap} \geq 0$ the last activation function of the actor is chosen as ReLU. In Table~\ref{tab:PPOpara}, all parameters used in the PPO algorithm are shown. 
In our trainings, we perform 2000 episodes each with 100 time steps. Each run through the system is followed by 100 successive training epochs.
\begin{table}[tb]
    \caption{Parameters used in the PPO algorithm.}
    \label{tab:PPOpara}
    \centering
    \begin{tabularx}{11.5cm}{p{7.0cm}|X} 
    \hline
    Parameter  &  Value \tabularnewline
    \hline
    \hline
    Hidden layers size actor and critic networks & [400, 300]  \\
    \hline
    Activation functions actor network  & [tanh, tanh, ReLU]  \\
    \hline
    Activation functions critic network  & [tanh, tanh, -ReLU]  \\
    \hline
    Actor's learning rate $\alpha_{a}$ & $5 \cdot 10^{-5}$  \\
    \hline
    Critic's learning rate $\alpha_{c}$ &  $1.5 \cdot 10^{-3} $ \\
    \hline
    Batch size $L$ & 100  \\
    \hline
    Reply buffer size  & 10000  \\
    \hline 
    Initial standard deviation stochastic actor & 10 \\
    \hline 
    c & 0.1 \\
    \hline 
    $\tau$ & 0.001 \\
    \hline 
    Entropy coefficient $\mu$ & 0.01 \\
    \hline 
    \end{tabularx}
\end{table}
\subsection{Simulation procedure}
\label{simu}
In the following, we will evaluate our approach on three different experiments. For each experiment 15 independent simulations are performed. In each simulation the algorithm is trained using 2000 different training references (one for each episode). After every tenth episode of the training the actor is tested on an evaluation reference. Using the evaluation results the best actor of the training is identified. In the next step, the best actor is applied to 100 test references to evaluate the performance of the algorithm. In the following, the mean episodic reward over all 100 test references and all 15 simulations  will serve as quality criterion. 

The smooth references are cubic splines formed from eight random values around the mean \SI{2000}{rpm}.
In the second experiment, between one and 19 discontinuities are induced by adding periodic square-waves of different period durations to the reference. The references in the third experiment are shifted by adding an offset to the spline. Five different offsets are used, resulting in data  between approximately \SI{1040}{rpm} and \SI{4880}{rpm}.

As mentioned before, the results will be compared to the performance of a PI controller. We determine the parameters of the PI controller by minimizing the cumulative costs over an episode, respectively the negative of the episodic reward defined by the reward function \eqref{eq:rewardCl}. To avoid local minima, we first apply a grid search and then use a quasi-Newton method 
to determine the exact minimum. In the optimization, the cumulative costs over all training references (same as in the PPO training) are used as quality criterion.
\subsection{Results}
\begin{figure}[tb]
    \centering
    \includegraphics[width=0.5\textwidth]{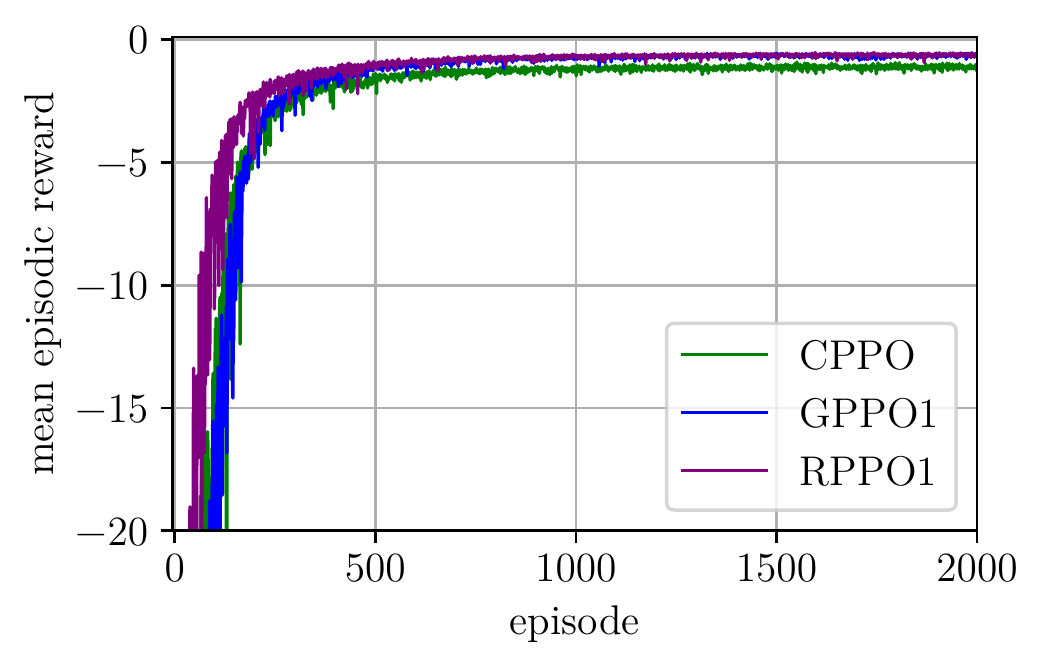}
    \caption{Training over 2000 episodes on smooth references.}
\label{fig:smoothTrain}
\end{figure}
\begin{table}[b]
    \caption{Quality evaluated on test a set for smooth references.}
    \label{tab:smoothTest}
    \centering
    \begin{tabularx}{15.0cm}{p{5.0cm}|p{2.0cm}|X|X} 
    \hline
    Arguments of actor and critic & Acronym & Mean episodic & Standard deviation \tabularnewline
      &  &  reward & episodic reward \tabularnewline
    \hline
    \hline
    Current state and residuum ($\vec{s}_k^{0,cl}$) & CPPO & -0.303 & 0.027 \\
    \hline
    Global space with one future reference ($\vec{s}_k^{11,cl}$) & GPPO1 & -0.035 & 0.020 \\
    \hline
    Residual space with one future reference ($\vec{s}_k^{21,cl}$) & RPPO1 & -0.030 & 0.020 \\
    \hline
    PI controller & PI & -0.069 & 1.534 \\
    \hline
    \end{tabularx}
\end{table}
\begin{figure}[bt] 
    \centering
    \subfloat[Input speeds and reference for one episode.] {\includegraphics[height=5.2cm]{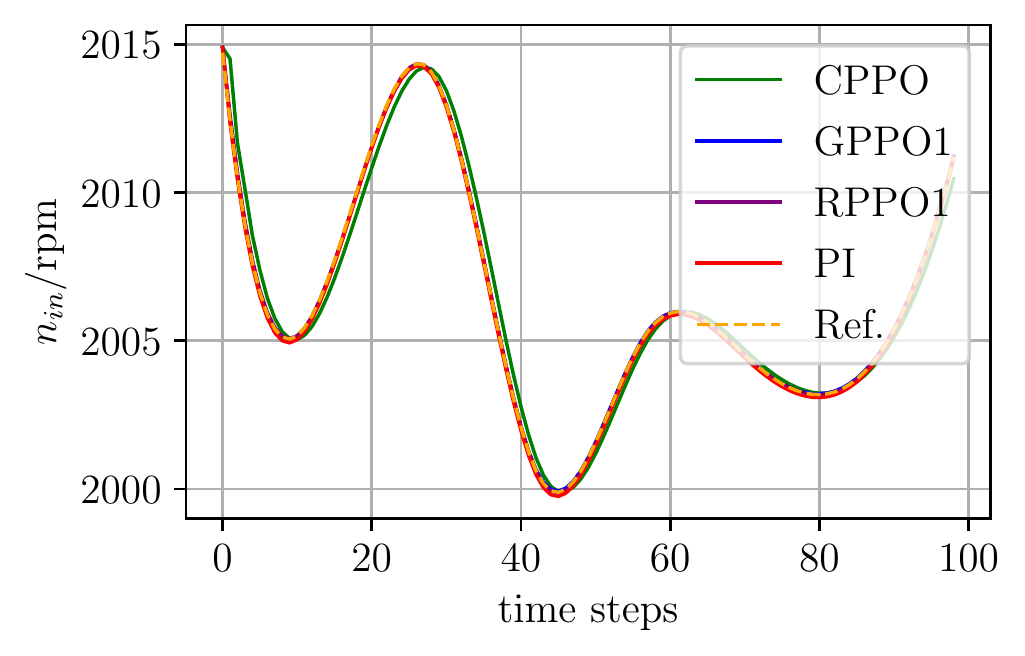}\label{fig:testStatesSmooth}} 
    \subfloat[Control inputs for one episode.]  {\includegraphics[height=5.2cm]{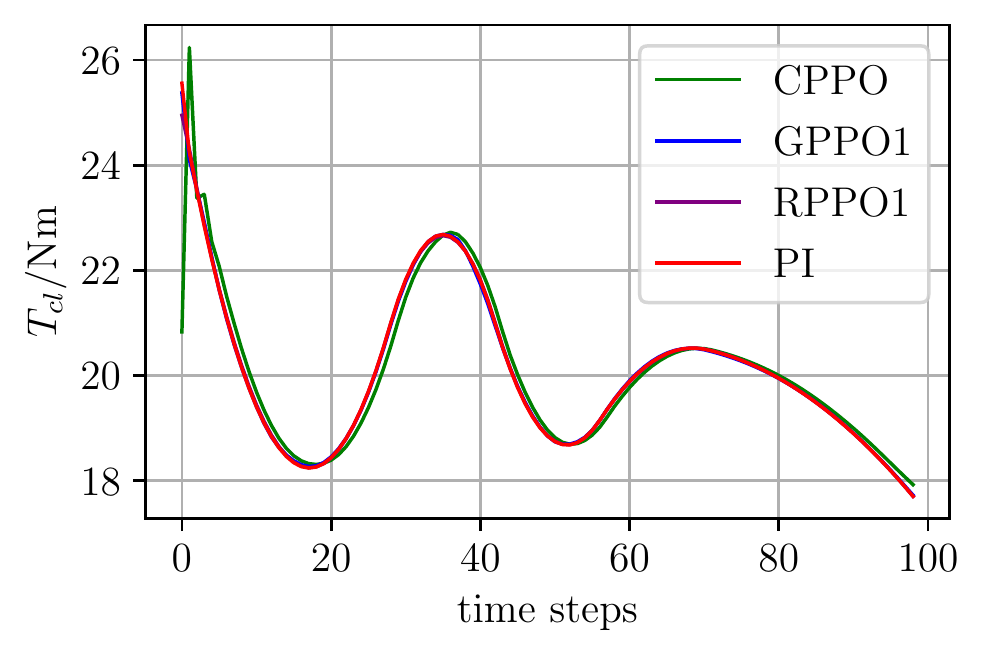}\label{fig:testActionssmooth}} 
    \caption{Performance of the best trained actors on a smooth test reference. Note that in real applications reference trajectories are designed in such a way that the input speed is synchronized quickly but smoothly towards zero slip at the clutch.}
    \label{fig:smoothRef}
\end{figure}
In the following, the results of the different experiments with (1) a class of smooth references, (2) a class of references with discontinuities and  (3) a class of smooth references shifted by an offset  will be presented.
\subsubsection*{Class of smooth references}
For smooth references the weighting parameter~$\beta$ in the reward function \eqref{eq:rewardCl} is set to zero. 
The parameters of the PI controller were computed as described in Section \ref{simu}, the parameter of the proportional term $K_P$ was determined as $20.88$ and the parameter of the integral term $K_I$  is $11.08$. The training curves of the PPO approaches are drawn in Figure~\ref{fig:smoothTrain}. It can be clearly seen that the approaches using one future reference value in the argument of the actor and the critic (GPPO1, RPPO1) reach a higher mean episodic reward than the approach including only reference information of the current time step (CPPO). Furthermore, the approach with the argument defined in the residual space (RPPO1) achieves high rewards faster than the GPPO1 and the CPPO.

As mentioned before, the trained actors are evaluated on a set of test references. The obtained results are given in Table \ref{tab:smoothTest}. The approach using global reference values in the argument (GPPO1) and the approach defining the argument in a residual space (RPPO1) provide the best results. The mean episodic reward of the CPPO is ten times lower than the ones of the GPPO1 and the RPPO1. The performance of the classical PI controller lies in between them but the standard deviation is very high. 
This implies that optimizing the PI's parameters using the training data, no parameters can be determined which lead to equally good performance for all the different references.

In Figure~\ref{fig:smoothRef}, the best actors are applied to follow a test reference. The tracks of the PI controller, the GPPO1 and the RPPO1 show similar behavior. Only in the valleys the PI controller deviated slightly more from the reference. 
Since the CPPO has to decide for a control input only knowing the current reference value, its reaction is always behind. In Figure~\ref{fig:testStatesSmooth}, it can be seen that the CPPO applies a control input that closes the gap between the current input speed and the current reference value. But in the next time step the reference value has already changed and is consequently not reached. The same effect leads to a shift of the control inputs in Figure~\ref{fig:testActionssmooth}. 
\subsubsection*{Class of references with discontinuities}
As mentioned before, to avoid unrealistic fast changes of the control inputs we add a first-order lag element with cutoff frequency $f_g = \SI{100}{Hz}$ to the traindrive model in the experiment setting for references with discontinuities. In addition, $\beta$ is set to $1/3000$ in the reward function \eqref{eq:rewardCl} to prevent the learning of large control inputs when discontinuities occur in the reference signal. The PI controller parameters were optimized on the same reward function resulting in $K_P=18.53$ and $K_I=5.67$.
\begin{figure}[tb]
    \centering
    \includegraphics[width=0.5\textwidth]{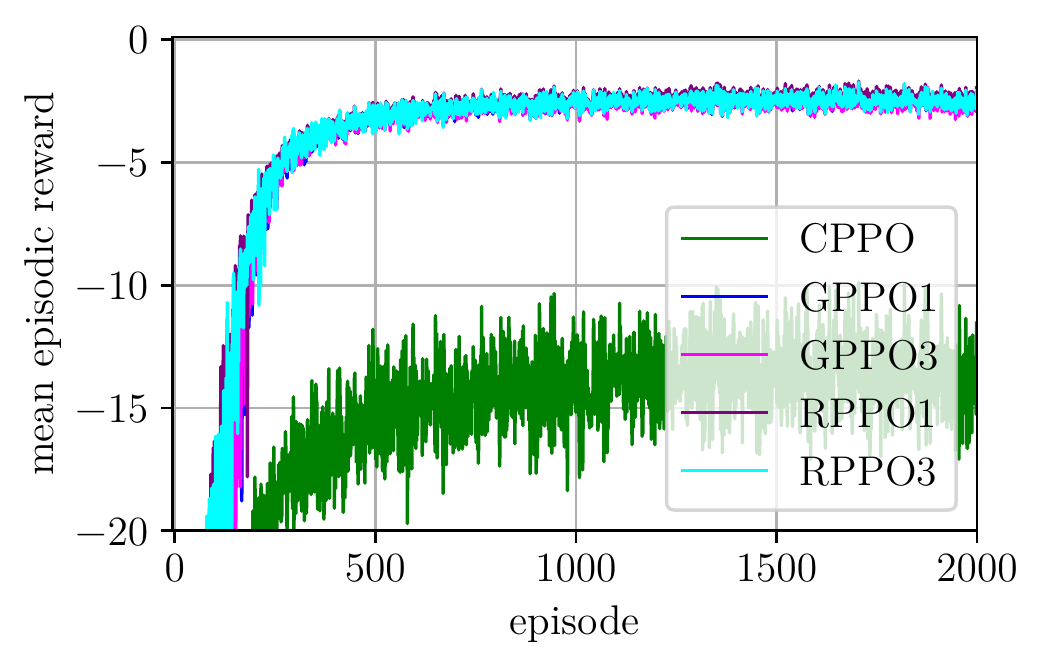}
    \caption{Training over 2000 episodes on references with discontinuities.}
\label{fig:jumpTraining}
\end{figure}
Besides including the information of one future reference value the performance of adding three future reference values (GPPO3 and RPPO3) is also investigated in this experiment. The learning curves of all five PPO approaches are illustrated in Figure~\ref{fig:jumpTraining}. All approaches with future reference values show similar results in the training. The PPO approach containing only current reference values (CPPO) achieves a significant lower mean episodic reward.

A similar behavior can be observed in the evaluation on the test reference set for which the results are given in Table~\ref{tab:jumpTest}.
\begin{table}[bt]
    \caption{Quality evaluated on test references with discontinuities.}
    \label{tab:jumpTest}
    \centering
    \begin{tabularx}{15.0cm}{p{5.0cm}|p{2.0cm}|X|X} 
    \hline
    Arguments of actor and critic & Acronym & Mean episodic & Standard deviation \tabularnewline
      &  &  reward & episodic reward \tabularnewline
    \hline
    \hline
    Current state and residuum ($\vec{s}_k^{0,cl}$) & CPPO & -10.936 & 0.618 \\
    \hline
    Global space with one future reference ($\vec{s}_k^{11,cl}$) & GPPO1 & -1.597 & 0.079 \\
    \hline
    Global space with three future reference ($\vec{s}_k^{13,cl}$)& GPPO3 & -1.499 & 0.082 \\
    \hline
    Residual space with one future reference ($\vec{s}_k^{21,cl}$) & RPPO1 & -1.614 & 0.088 \\
    \hline
    Residual space with three future reference ($\vec{s}_k^{23,cl}$) & RPPO3 & -1.510 & 0.086 \\
    \hline
     PI controller & PI & -1.747 & 2.767 \\
    \hline
    \end{tabularx}
\end{table}
Again, the PPO approaches with one global future reference (GPPO1) and one future reference in the residual space (RPPO1) show similar results and perform significantly better than the CPPO. Adding the information of more future reference values (GPPO3 and RPPO3) leads to an even higher mean episodic reward. The PI controller performs slightly worse than the PPO aproaches with future reference information. Due to the large standard deviation of the PI controller's performance, it cannot be guaranteed to perform well on a specific reference.

\begin{figure}[b] 
    \centering
    \subfloat[Input speeds and reference.]{\includegraphics[height=5.2cm]{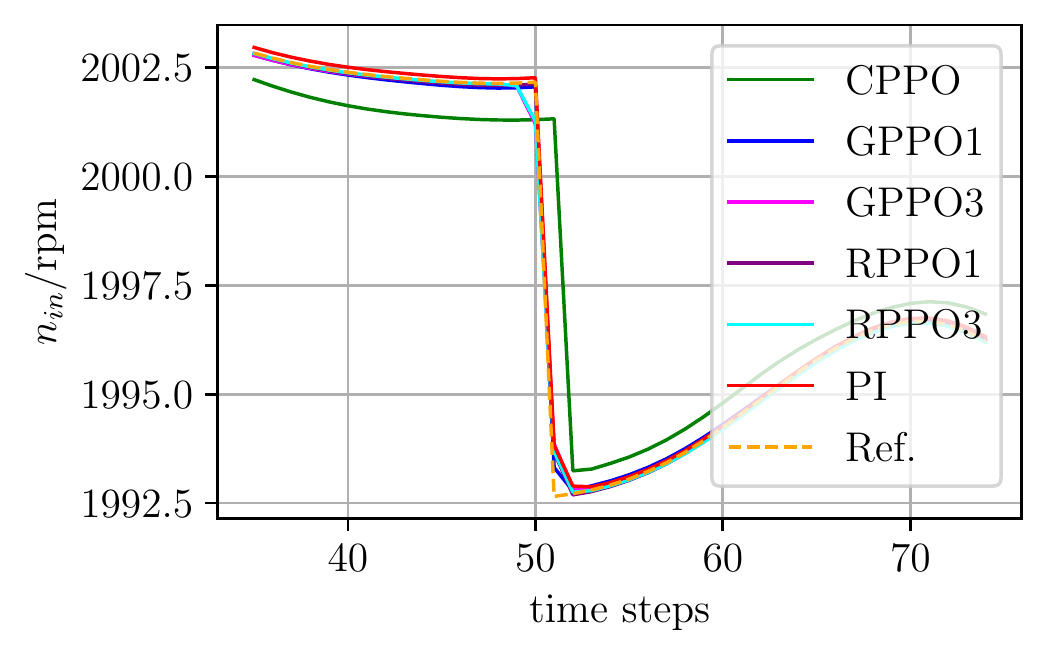}\label{fig:testStatesjumps}} 
    \subfloat[Control inputs.]{\includegraphics[height=5.2cm]{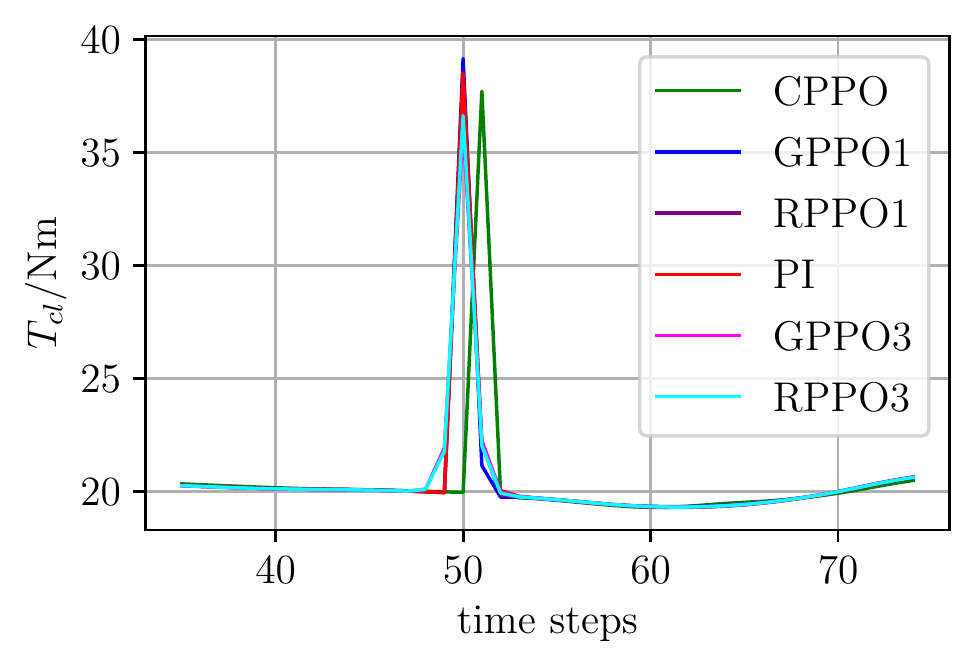}\label{fig:testActionsjumps}} 
    \caption{Performance of the best trained actors on a test reference with discontinuities.}
\label{fig:testjumps}
\end{figure}

The performance of the best actors on a test reference with discontinuities is illustrated in Figure~\ref{fig:testStatesjumps}. Note, for better visibility, the figure shows only a section of a full episode. 
The PPO approaches including future reference values and the PI controller are able to follow the reference even if a discontinuity occurs. The PPO approach without future reference values is performing even worse than on the smooth reference. It can be clearly seen that the jump cannot be detected in advance and the reaction of the actor is delayed. 
In Figure~\ref{fig:testStatesjumps} the applied control inputs are shown, the approaches including three future reference values respond earlier to the upcoming discontinuities and cope with the discontinuity requiring only smaller control inputs (Figure~\ref{fig:testActionsjumps}) which also leads to a higher reward. The PI controller shows a similar behavior than the PPO approaches with one future reference value (GPPO1 and RPPO1).
\subsubsection*{Class of smooth references with offsets}
\begin{figure}[tb]
    \centering
    \includegraphics[width=0.5\textwidth]{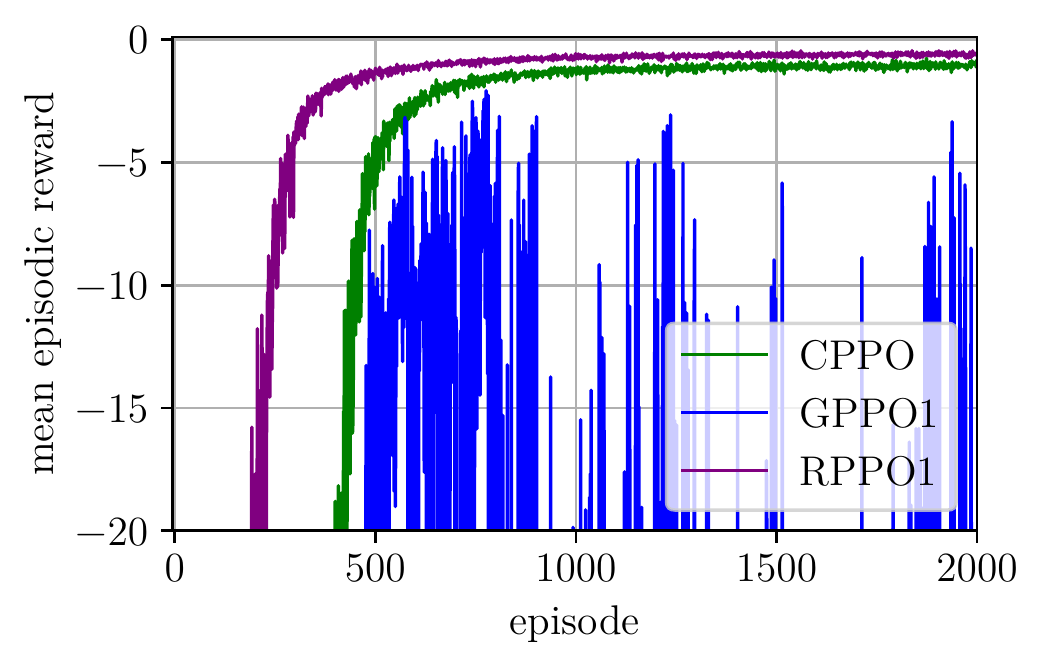}
    \caption{Training over 2000 episodes on smooth references with offsets.}
    \label{fig:offsetTrain}
\end{figure}

For the class of smooth references with offsets, the same settings, as used in the experiment for smooth references, are applied. The PI parameters are determined as $20.88$ for $K_P$ and $11.04$ for $K_I$. The training curves of the PPO approaches are illustrated in Figure~\ref{fig:offsetTrain}.
Compared to the learning curves of the smooth references experiment the training on the references with offset needs more episodes to reach the maximum episodic reward. The training of the approach using global reference values (GPPO1) is extremely unstable. The reason might be that the training data contains only references in a finite number of small disjointed areas in a huge state space. As the argument of GPPO1 includes only global states and reference value no shared data between this areas exist and the policy is learned independently for each area.
Consequently, the stability of the approaches with argument components in the residual space CPPO and RPPO is not effected by this issue. However, the RPPO1 is receiving higher rewards significantly faster than the CPPO which also contains the global input speed. 
\begin{table}[b]
    \caption{Quality evaluated on a test set for smooth references with offset.}
    \label{tab:offsetTest}
    \centering
    \begin{tabularx}{15.0cm}{p{5.0cm}|p{2.0cm}|X|X} 
    \hline
    Arguments of actor and critic & Acronym & Mean episodic & Standard deviation \tabularnewline
      &  &  reward & episodic reward \tabularnewline
    \hline
    \hline
    Current state and residuum ($\vec{s}_k^{0,cl}$) & CPPO & -0.377 & 0.091 \\
    \hline
    Global space with one future reference ($\vec{s}_k^{11,cl}$) & GPPO1 & -0.102 & 0.089 \\
    \hline
    Residual space with one future reference ($\vec{s}_k^{21,cl}$) & RPPO1 & -0.031 & 0.018 \\
    \hline
    PI controller & PI & -0.069 & 1.541 \\
    \hline
    \end{tabularx}
\end{table}

In Table \ref{tab:offsetTest}, the performance on a set of test references is given. The mean episodic reward of the CPPO, the RPPO1  and the PI controller is in the same range as in the smooth references experiment. As in the other experiments, the PI controller shows a large standard deviation. Despite the instabilities in the training, the GPPO1 receives a higher mean episodic reward than the CPPO but its performance is significant worse than in the smooth references experiment. It can be clearly seen that residual space arguments lead to better performance for references with offsets.
\begin{figure}[t] 
    \centering
    \subfloat[Input speeds and reference.]{\includegraphics[height=5.2cm]{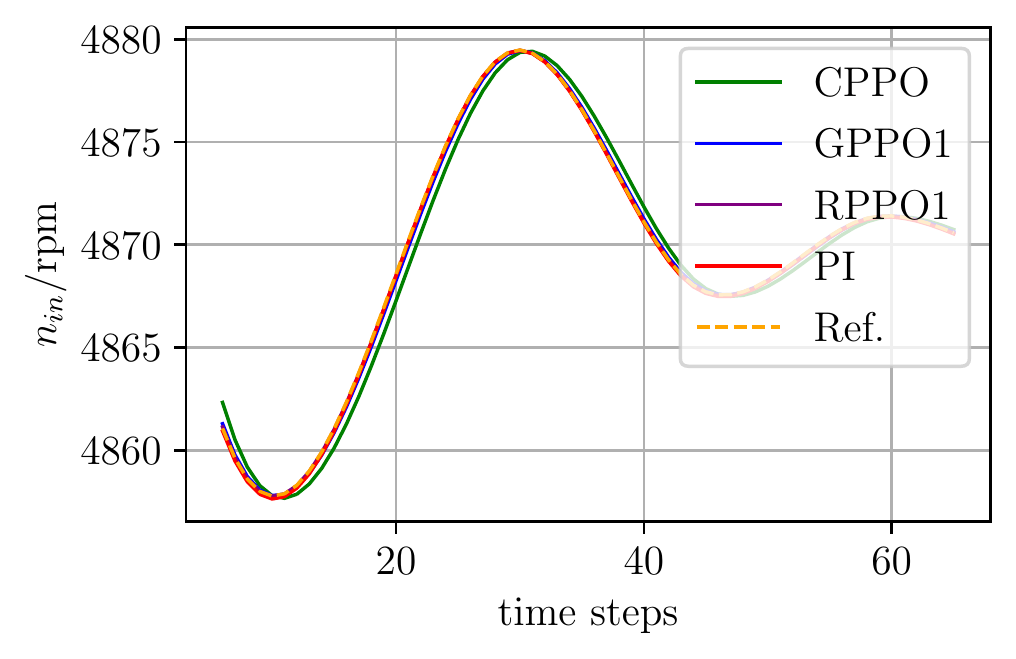}\label{fig:offsetStates}} 
    \subfloat[Control inputs.]{\includegraphics[height=5.2cm]{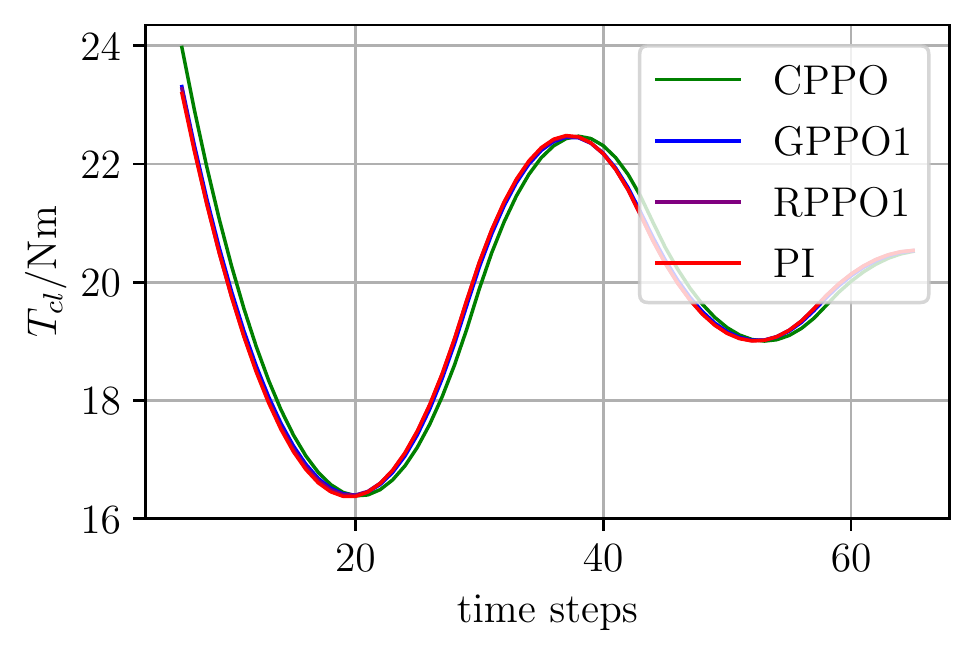}\label{fig:offsetActions}} 
    \caption{Performance of the best trained actors on a smooth test reference with offsets.}
    \label{fig:offsetPlots}
\end{figure}
In Figure~\ref{fig:offsetPlots}, the performance of the best actors on a test reference is illustrated. For better visibility, only a section of a full episode is drawn. 
The results are similar to the smooth reference experiment, only the GPPO1 shows slightly worse tracking performance.

\section{Conclusion}

In this work, proximal policy optimization for tracking control exploiting future reference information was presented. We introduced two variants of extending the argument of both actor and critic. In the first variant, we added global future reference values to the argument. In the second variant, the argument was defined in a novel kind of residual space between the current state and the future reference values. By evaluating our approach on a simple drive train model we could clearly show that both variants improve the performance compared to an argument taking only the current reference value into account. If the approach is applied to references with discontinuities, adding several future reference values to the argument is beneficial. The residual space variant shows its advantages especially for references with different offsets. In addition, our approach outperforms PI controllers commonly used in drive train control. Besides higher tracking quality, the generalization to different references is significantly better than using a PI controller. This guarantees an adequate performance on arbitrary, before unseen, references. 

In future work, our approach will be applied to a more sophisticated drive train model where noise behavior, model inaccuracies, non-modeled dynamics, and control input range limits, as being expected in a real drive train system, are systematically included in training. 


\bibliographystyle{IEEEtran}
\bibliography{refer}

\end{document}